\documentclass[conference]{IEEEtran}
\IEEEoverridecommandlockouts                              

\usepackage{graphicx}
\usepackage[letterpaper, left=0.75in, right=0.75in, bottom=0.8in, top=0.75in]{geometry}

\usepackage{amsmath} 
\usepackage{amssymb}  
\usepackage{mathrsfs}
\usepackage{todonotes}
\usepackage{enumerate}
\usepackage{float}
\usepackage{caption}
\usepackage{subcaption}
\usepackage{cite}
\usepackage{tabularx}
\usepackage[ruled,vlined,linesnumbered]{algorithm2e}
\usepackage{bm}
\usepackage{xcolor}

\newtheorem{definition}{Definition}
\newtheorem{proposition}[definition]{Proposition}

\title{\LARGE \bf
Experiments in Underwater Feature Tracking with Performance Guarantees Using a Small AUV
}

\author{\IEEEauthorblockN{Benjamin Biggs}
\IEEEauthorblockA{\textit{Virginia Tech}\\
babiggs@vt.edu}
\and
\IEEEauthorblockN{Hans He}
\IEEEauthorblockA{\textit{Virginia Tech}\\
hjh2bs@vt.edu}
\and
\IEEEauthorblockN{James McMahon}
\IEEEauthorblockA{\textit{US Naval Research Laboratory}\\ Acoustics Division, Code 7130 \\
    james.mcmahon@nrl.navy.mil}
\and 
\IEEEauthorblockN{Daniel J. Stilwell}
\IEEEauthorblockA{\textit{Virginia Tech}\\
stilwell@vt.edu}
\thanks{*This work was supported by the Office of Naval Research via grants N00014-18-1-2627, and N00014-19-1-2194. The work of J. McMahon is supported by the Office of Naval Research through the NRL Base Program.}
\thanks{James McMahon is with the US Naval Research Laboratory, Code 7130, Washington D.C., USA}%
\thanks{Benjamin Biggs, Hans He, and Daniel Stilwell are with the Bradley Department of Electrical and Computer Engineering, Virginia Tech, Blacksburg, VA, USA}%

}

\begin{document}
\maketitle
\thispagestyle{empty}
\pagestyle{empty}

\begin{abstract}
    We present the results of experiments performed using a small autonomous underwater vehicle to determine the location of an isobath within a bounded area.  The primary contribution of this work is to implement and integrate several recent developments real-time planning for environmental mapping, and to demonstrate their utility in a challenging practical example.   We model the bathymetry within the operational area using a Gaussian process and propose a reward function that represents the task of mapping a desired isobath. As is common in applications where plans must be continually updated based on real-time sensor measurements, we adopt a receding horizon framework where the vehicle continually computes near-optimal paths. The sequence of paths does not, in general, inherit the optimality properties of each individual path. Our real-time planning implementation incorporates recent results that lead to performance guarantees for receding-horizon planning.   
\end{abstract}

\section{INTRODUCTION}


Determining the location of a depth contour or isobath is an important element of quickly finding navigable paths in waterways. The conventional approach to underwater feature mapping relies on performing an exhaustive search over an area where a human operator believes features of interest are located. Path planning methods that seek to maximize information gain by planning sequences of locations that maximize a specific reward function are available to attempt to track underwater features more quickly than exhaustive search methods. However, many of these methods are susceptible to poor performance caused by myopic planning horizons when using receding horizon methods or local maxima in the reward function when using gradient-based optimization approaches. The primary contribution of this paper is the presentation of the results of experiments in depth contour estimation using path planning techniques with performance guarantees. We integrate  several recent developments for real-time isobath localization, all of which can be easily modified for similar real-time mapping applications: (1) use of a sparse Gaussian process for representing the bathymetry, which leads to a continuous representation of the environment along with rigorous assessment of uncertainty, and (2) use of a so-called terminal reward in the optimization problem that generates each near-optimal path, which ensures that the cumulative reward attained by the AUV from a sequence of near-optimal paths is no worse than a desired lower bound.   An important contribution of this work is to describe specific implementation solutions needed to integrate and implement these ideas for real-time operation on an AUV.

\begin{figure}
\centering
\includegraphics[width=0.98\columnwidth]{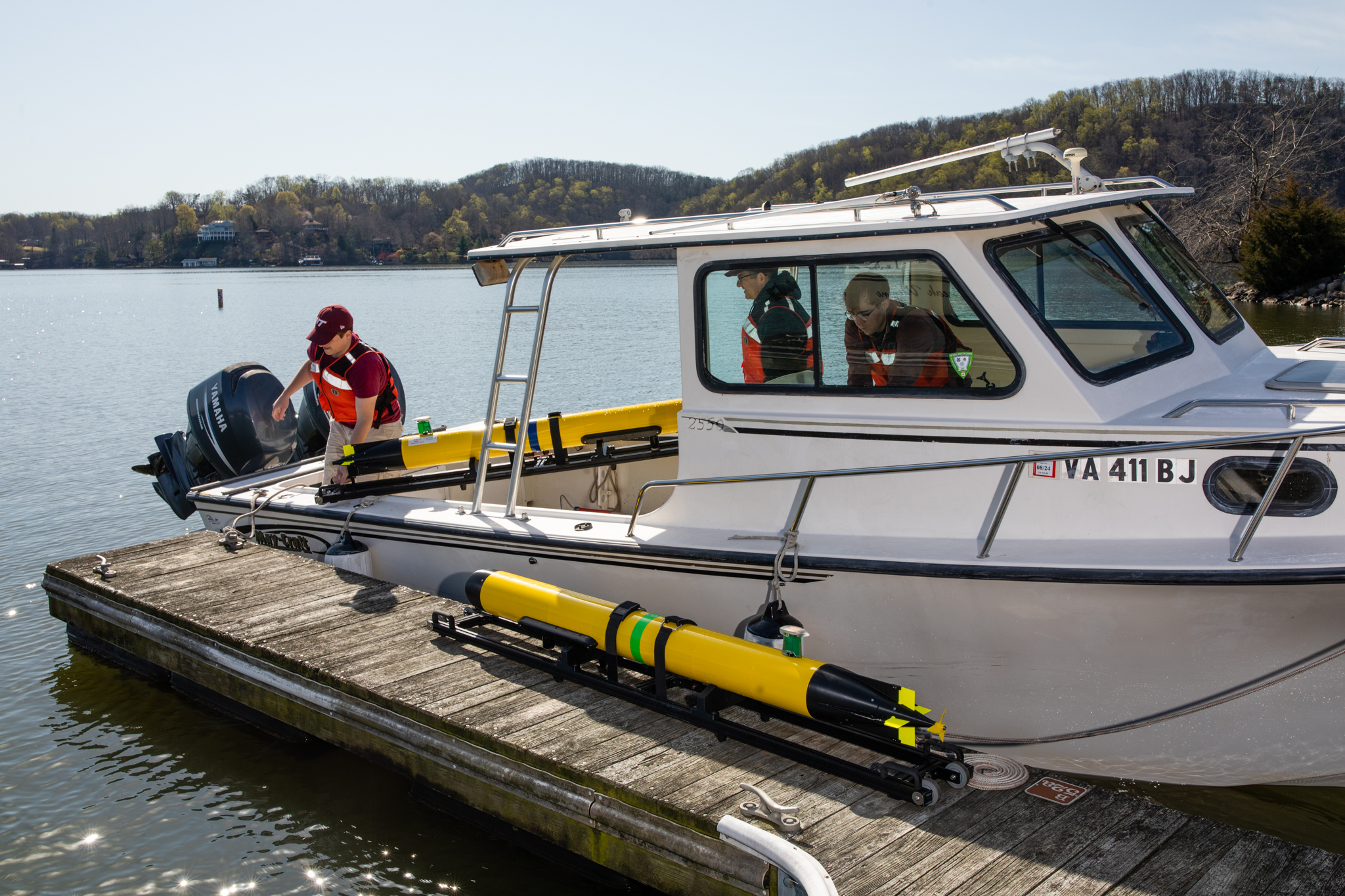}
\caption{690 vehicles being loaded for transportation to the operational area.}
\label{fig.690}
\end{figure}

Isobath localization, like front tracking, can be formulated as an informative path planning (IPP) problem \cite{mccammon2021ocean}. The goal of the IPP problem is to find a path that maximizes an information utility function while not exceeding a budget constraint on the cost of the path. In general, the computational complexity associated with informative path planning grows exponentially with the length of the path. Heuristic planning methods may be used to produce reasonable results \cite{frolov2014can, zhang2016autonomous, dunbabin2017quantifying}. However, it is desirable to plan paths that are as nearly optimal as possible. Despite a host of techniques developed to provide approximate or suboptimal solutions \cite{singh2009nonmyopic, schlotfeldt2018anytime, yetkiny2019online, kantaros2019asymptotically}, receding horizon methods are often used regardless of the planning techniques used within the shorter planning horizon \cite{atanasov2014information, atanasov2015decentralized, wakulicz2022informative}.

A receding horizon path is constructed from segments of near-optimal short paths that are greedily selected. However, the sequence of path segments in a receding horizon implementation do not inherit the near-optimality of each individual segment. In prior work \cite{biggs2019performance, biggs2020extended} the authors showed how to append a specific terminal reward to the short horizon optimization problem used for each path segment such that the reward of the overall receding horizon path is guaranteed to exceed a desirable lower bound. This result was demonstrated for the explicit reward function associated with a robotic search problem in a discrete environment through numerical experiments. In this work, we demonstrate the utility of adding a terminal reward to a different class of reward function based on level set estimation using Gaussian process models. We report on the results of challenging real-world experiments accomplished using a real autonomous underwater vehicle (AUV) performing on-line planning using live data.

Throughout this work, we seek to rapidly map the 15 meter isobath within a bounded area of Claytor lake in Dublin, VA, USA. The specific region of Claytor lake where the experiments of this work are performed is shown in Figure \ref{fig.search_area} with the solid red lines indicating the boundaries of the operational area. The AUV used for these experiments is the Virginia Tech 690 vehicle shown in Figure \ref{fig.690}.


The remainder of this paper is organized as follows. Gaussian process regression and the specific considerations and methods used in this work are presented in Section \ref{sec.gpr}. The specific reward function used for evaluating paths is presented in Section \ref{sec.obj_func}. The path planning methods proposed in \cite{biggs2020extended} and used in this work are presented and discussed in Section \ref{sec.rh}. The usage and relevant capabilities of the 690 vehicle are discussed in Section \ref{sec.690}. The experimental procedure and selection of parameters is presented in Section \ref{sec.trials}. Results and conclusions are presented in Section \ref{sec.conclusions}.

\section{GAUSSIAN PROCESS REGRESSION}\label{sec.gpr}
In this section we provide a brief review of Gaussian process regression. We refer the reader to \cite[Chapter~2]{williams2006gaussian} for a more thorough discussion of the topic.

A Gaussian process is a collection of random variables, any finite number of which have a joint Gaussian distribution \cite[Definition~2.1]{williams2006gaussian}. A GP is completely specified by its mean and covariance functions $m(p)$ and $k(p, p')$ respectively and is written
\begin{equation}
    f(p) \sim \mathcal{GP}(m(p), k(p, p'))
\end{equation}
where $f(p)$ is a real process. In this work, we assume that the bathymetry or depth of Claytor lake within the operational area may be modeled using a GP. 

A sample or datum $d = (p,z)$ is an input-measurement pair with $p \in \mathscr{P}$ and $z \in \mathbb{R}$. Let $\mathcal{D}_t$ be the set of all samples acquired up to time $t$. For any input $p_i$, an associated measurement $z_i$ is related to $p_i$ via
\begin{equation}
    z_i = f(p_i) + \epsilon_i, \ \text{for} \ i = 1, \ldots, |\mathcal{D}_t|
\end{equation}
where $|\mathcal{D}|$ denotes the cardinality of the set $\mathcal{D}$ and $\epsilon_i \sim \mathcal{N}(0, \sigma_n^2)$ is drawn from a zero mean Gaussian distribution with standard deviation $\sigma_n$. 

Given the data set $\mathcal{D}_t$, one may wish to predict $f(p^*)$ for an arbitrary input $p^*$. From \cite[Chapter~2.7]{williams2006gaussian}, the posterior predictive distribution of $f(p^*)$ conditioned on $\mathcal{D}_t$ is Gaussian and specified by the mean $\mu_{p^*| \mathcal{D}_t}$ and variance $\sigma^2_{p^*| \mathcal{D}_t}$ given as
\begin{align} \label{eq.pred_mean}
    \mu_{p^*| \mathcal{D}_t} &= m(p^*) + k(p^*)^\top(K_t + \sigma_n^2I)^{-1}(Z_t - m(P_t)) \\
                \label{eq.pred_var}
    \sigma^2_{p^*| \mathcal{D}_t} &= k(p^*, p^*) - k(p^*)^\top(K_t + \sigma_n^2I)^{-1}k(p^*).
\end{align}
$Z_t = [z_1, \ldots, z_{|\mathcal{D}_t|}]^\top$ is the vector of all measurements contained in $\mathcal{D}_t$ and $P_t = \{p_1, \ldots, p_{ |\mathcal{D}_t|}\}$ is the set of associated inputs. The prior mean vector is $[m(P_t)]_i = m(p_i)$, $[k(p^*)]_i = k(p^*, p_i)$, $[K_t]_{i,j} = k(p_i, p_j)$, and $I_t$ is the $|\mathcal{D}_t| \times |\mathcal{D}_t|$ identity matrix.

When computing the predictive distribution at $p^*$, the matrix inversion $(K_t + \sigma_n^2I)^{-1}$ can become intractable as $|\mathcal{D}_t|$ grows large. To address this complexity, we utilize a distance metric $d(p, p')$ on $\mathscr{P}$ to reduce the number of samples retained in $\mathcal{D}_t$ and make predictions using local Gaussian processes. To reduce the number of samples retained in $\mathcal{D}_t$, a new sample $d_c = (p_c, z_c)$ is only added to $\mathcal{D}_t$ if $d(p_c, p_i) \geq \delta_f$ for all $p_i \in \mathcal{D}_t$. For a predictive point $p^*$, the associated data is limited to a locality around $p^*$ giving the local data set as $\mathcal{D}_{p^*,t} = \{ d_i \in \mathcal{D}_t | d(p_i, p^*) \leq \delta_c\}$. The local predictive mean and variance are given by equations \eqref{eq.pred_mean} and \eqref{eq.pred_var} using $\mathcal{D}_{p^*,t}$ in the place of $\mathcal{D}_t$.

Throughout this work, we define the input space $\mathscr{P}$ to be two dimensional euclidean space ($\mathbb{R}^2$) with distance metric $d(p, p') = \| p - p' \|$ being the euclidean distance. We use the squared exponential kernel function
\begin{equation}\label{eq.se}
    k(p, p') = \sigma_f^2 \exp \left\{ \frac{\| p - p' \|^2}{2l_c^2} \right\}
\end{equation}
as the prior for the covariance function. The prior mean function used in this work is presented in Section \ref{sec.trials}.

\section{OBJECTIVE FUNCTION}\label{sec.obj_func}
We wish to determine the level set $\mathcal{L}(l^*) = \{ p \in \mathscr{P} | f(p) = l^*\}$. To direct the selection of locations that should be measured in order to obtain information regarding this level set, we require an objective function that encodes uncertainty regarding whether $p^*$ is an element of $\mathcal{L}(l^*)$. To provide such a function, we use a measure of ambiguity presented in \cite{gotovos2013active} which is closely related to the straddle heuristic presented in \cite{bryan2005active}.
\begin{definition}[Ambiguity] \label{def.amb}
    The ambiguity associated with a point $p^*$ with respect to a level $l$ is
    \begin{equation}\label{eq.ambiguity}
        a_{p^*|\mathcal{D}_t} = \beta \sigma_{p^*| \mathcal{D}_t} - |\mu_{p^*| \mathcal{D}_t} - l|.
    \end{equation}
\end{definition}
Concretely, if $\beta = 2$ and $a_{p^*|\mathcal{D}_t} < 0$, then the model indicates that the probability that $f(p^*)$ falls on the opposite side of $l$ as $\mu_{p^*| \mathcal{D}_t}$ is less than 2.5 percent.

Similar to \cite{gotovos2013active} where batches of points are selected for sampling, we seek to select a path or a sequence of points to sample. Considering the ambiguity of sets of points requires some slight modifications to account for locations that will be sampled while no samples have yet been acquired.
\begin{definition}[Anticipated Ambiguity]
    Let $\mathcal{P}^* = \{p_0^*, \ldots, p_{n}^*\}$ and $\mathcal{P}_i^* = \{p_0^*, \ldots, p_{i-1}^*\}$ with $\mathcal{P}_0^* = \emptyset$. Then the anticipated ambiguity of $p_i^* \in \mathcal{P}^*$, with respect to a level $l$ is
    \begin{equation}
        a'_{p_i^*|\mathcal{D}_{t}, \mathcal{P}_i^*} = \beta \sigma_{p_i^*| \mathcal{D}_t, \mathcal{P}_i^*} - |\mu_{p^*| \mathcal{D}_t} - l|
    \end{equation}
    where $\sigma_{p_i^*| \mathcal{D}_t, \mathcal{P}_i^*}$ is the posterior predictive standard deviation computed using equation \eqref{eq.pred_var} where $\mathcal{P}_t$ is augmented with $\mathcal{P}_i^*$.
\end{definition}

Having defined the anticipated ambiguity for a set of points, we define our reward function to be the anticipated reduction in ambiguity at a point with respect to a level $l$.
\begin{equation}\label{eq.reward}
    r_{p_i^*|\mathcal{D}_{t}, \mathcal{P}_i^*} = \max(a'_{p_i^*|\mathcal{D}_{t}, \mathcal{P}_i^*}, 0) - \max(a'_{p_i^*|\mathcal{D}_{t}, \mathcal{P}_i^* \cup p_i^*}, 0).
\end{equation}
Note that $a'_{p_i^*|\mathcal{D}_{t}, \mathcal{P}_0^*} = a_{p_i^*|\mathcal{D}_{t}}$. Also note that without the inclusion of the $\max$ function, the reward simply becomes $\beta [\sigma_{p_i^*| \mathcal{D}_t, \mathcal{P}_i^*} - \sigma_{p_i^*| \mathcal{D}_t, \mathcal{P}_i^* \cup p_i^*} ]$, which is uninformative regarding $\mathcal{L}(l)$. Intuitively, the reward places the most value on points $p^*$ where $f(p^*)$ is expected to be close to $l$ and the anticipated standard deviation at $p^*$ is expected to be reduced significantly. That is, $f(p^*)$ may be close to $l$, but if the anticipated reduction in variance is low, there is little to be gained by sampling that location.

\section{RECEDING HORIZON PATH PLANNING}\label{sec.rh}
As in \cite{biggs2020extended}, we consider that the state of a mobile search agent may be represented by the tuple $s = \{p, x \}$ where $x$ gives all states of the mobile search agent excluding the position states $p$. We say that a state $s_i$ is feasible from the state $s_{i-1}$ if there exists an action $a$ available to the search agent such that applying $a$ to the state $s_{i-1}$ results in the new state $s_i$. An $n$-length path may then be represented as a sequence of states $\gamma_n(s_0) = \{ s_0, s_1, \ldots, s_n \}$. The set of all feasible $n$-length paths beginning at state $s_0$ is denoted $\Gamma_n(s_0)$.

Using the point-wise reduction in ambiguity in  equation \eqref{eq.reward} as the reward function, the value of a path is the sum of rewards obtained while traversing the path. That is, the anticipated reward for the path $\gamma_n(s_k)$ is
\begin{equation}\label{eq.path_reward}
    J(\gamma_n(s_0)) = \sum_{i = 0}^n r_{p_i|\mathcal{D}_{t} \cup \mathcal{P}_i}
\end{equation}
where $p_i \in s_i \in \gamma_n(s_k)$ is the point to be sampled and $\mathcal{P}_i = \{p_0, \ldots, p_{i-1}\}$ is the set of points that will have been sampled by step $i$. For notational simplicity we define
\begin{equation}
    J_n(s_0) \triangleq J(\gamma_n(s_0)).
\end{equation}

The goal of path planning is to determine a path $\gamma_n^*(s_0)$ satisfying
\begin{equation}\label{eq.path_optimization_problem}
    \gamma_n^*(s_0) = \underset{\gamma_n(s_0) \in \Gamma_n(s_0)}{\arg \max}J_n(s_0).
\end{equation}
In general, the complexity of \eqref{eq.path_optimization_problem} grows exponentially as the length of the desired path increases. One method of addressing the complexity associated with planning long paths is to employ receding horizon planning methods wherein a search agent solves for a short optimal or nearly optimal path, traverses a portion of the short path, and repeats the process until a path of the desired length has been traversed. The length of the computationally feasible short paths is termed the \textit{planning horizon}. The number of steps that the agent traverses along each short path before planning a new short path is termed the \textit{execution horizon}.

While receding horizon path planning is commonly used in practice and generally yields good results, there are no optimality guarantees for paths produced using receding horizon methods despite the receding horizon paths being constructed from segments of optimal or nearly optimal short paths. In fact, receding horizon planning methods may produce worst possible solutions.

In prior works \cite{biggs2019performance} and \cite{biggs2020extended}, it was shown that the appropriate use of a terminal reward when constructing receding horizon paths can guarantee a lower bound on the value of receding horizon paths produced. The primary tool in producing this lower bound is a lower bound on the \textit{value-to-go} or a lower bound on the optimal value of the remainder of the path normally ignored during receding horizon planning.
\begin{definition}{\cite[Definition~3.1]{biggs2020extended}}[value-to-go]\label{def.vtg} A lower bound on the \textit{value-to-go} is the function $B_l(s_n)$ that satisfies
\begin{equation}
    B_l(s_n) \leq J_{l-n}^*(s_n)
\end{equation}
for all $n \leq l$ with $B_l(s_l) = 0$, and for the infinite case where $l = \infty$, $B_l(s_k) \rightarrow \infty$ as $k \rightarrow \infty$.
\end{definition}

Using the value-to-go function $B$, the authors of \cite{biggs2020extended} propose a new objective function as
\begin{equation}\label{eq.vn}
    \mathcal{V}(\gamma_n(s_0)) = J_n(s_0) + B_l(s_n)
\end{equation}
with the abbreviated notation
\begin{equation}
    \mathcal{V}_n(s_0) \triangleq \mathcal{V}(\gamma_n(s_0)).
\end{equation}
Using the objective function of equation \ref{eq.vn}, the lower bound guarantee that we seek leverage in the field trials of this work, is given in Proposition \ref{prp_better_than_prev}.
\begin{proposition}{\cite[Proposition~4.2]{biggs2020extended}}\label{prp_better_than_prev}
Suppose that at each planning step $k$ the path $\gamma_n(s_k)$ satisfies
\begin{equation}\label{eq.prop_2_hyp}
    \mathcal{V}_{n}(s_k) \geq \mathcal{V}_{n-m}(s_k)
\end{equation}
where $\mathcal{V}_{n-m}(s_k)$ is the value of the $n-m$ remainder of the path produced at the previous planning step and
\begin{equation}\label{eq.prop_2_hyp_2}
    \mathcal{V}_{n}(s_0) \geq B_l(s_0)
\end{equation}
then the $l$-length RH path $\tilde{\gamma}_l(s_0)$ satisfies
\begin{equation}
   J(\tilde{\gamma}_l(s_0)) \geq B_l(s_0).
\end{equation}
\end{proposition}
Significantly, it is shown in Corollary 4.3 of \cite{biggs2020extended} that the lower bound on the value-to-go may be defined using the value of the highest-reward path from a set of naive paths. This allows the receding horizon planner to leverage the strengths of distinct naive paths to produce a receding horizon path that will perform at least as well as the best naive solution and possibly much better.

\section{VIRGINIA TECH 690 AUV}\label{sec.690}
\begin{figure}
\centering
\includegraphics[width=0.8\columnwidth]{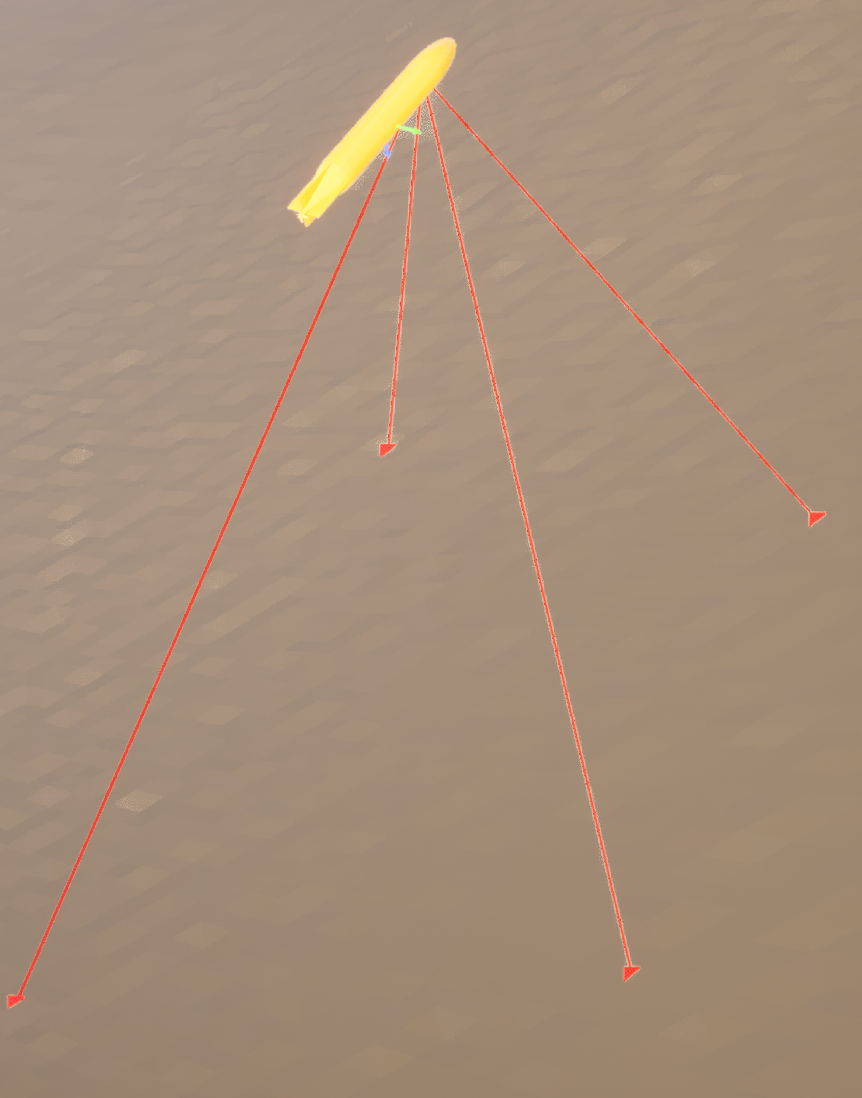}
\caption{Simulated DVL beams shown in red.}
\label{fig.dvl}
\end{figure}

Experiments were performed using the Virginia Tech 690 AUV shown in Figure \ref{fig.690}. Specifications for the 690 are given in Table \ref{tab.690_table}. Significantly, the 690 vehicle is equipped with a Doppler velocity log (DVL, TELEDYNE MARINE Pathfinder) with four acoustic beams at a 30 degree janus angle. An example AUV illustrating the DVL placement on the 690 is shown in Figure \ref{fig.dvl} with DVL beams shown in red. The DVL records a measurement of the distance from the DVL sensor to the lake bed along each beam at a rate of 10Hz. The 690 vehicle is able to estimate its location using DVL-aided inertial navigation. The navigation estimate is fused with the DVL range measurements to infer the depth of the lake bed. We use all four range measurements in our implementation. Throughout this paper, vehicle state, and sensor measurements are converted into a North, East, Down (NED) coordinate frame. The origin of the NED coordinate system is shown as the green point in Figure \ref{fig.search_area}.

\begin{table}[h]
\begin{tabular}{|m{0.45\columnwidth}|m{0.45\columnwidth}|}
    \hline
    \textbf{PARAMETER} & \textbf{SPECIFICATION} \\
    \hline
    Length & 2.2 m \\
    \hline
    Displacement & 42.9 kg \\
    \hline
    Diameter & 17.5 cm \\
    \hline
    Endurance & 24 hrs @ 2 m/s \\
    \hline
    Depth & 500 meters\\
    \hline
    Communications & 900MHz RF modem, Wi-Fi, acoustic communication, satellite communication \\
    \hline
    Navigation & DVL + IMU, acoustic ranging \\
    \hline
    Mission Sensor & PingDSP 3D bathymetric sonar \\
    \hline
\end{tabular}
\caption{Specifications of the Virginia Tech 690 AUV.}
\label{tab.690_table}
\end{table}

Control and navigation of the 690 vehicle are accomplished using ROS-based autonomy software running on an ODROID-XU4 single board computer.
Path planning for these experiments is accomplished separately on an NVIDA Jetson TX2. Line segments connecting waypoints in the paths computed onboard the Jetson are delivered to the ROS-based autonomy software on the ODROID via a TCP interface. The autonomy software running on the ODROID drives the 690 vehicle to converge to the input line segment and reports when the end of the line segment has been reached. DVL sensor measurements and vehicle navigation data are transmitted from the ODROID to the Jetson over the same interface. Notably, executing path planning on a separate computer and sending line segments over a well-defined interface, ensures that the path planning algorithm does not interfere with critical autonomy functionality. 

\section{FIELD TRIALS}\label{sec.trials}

\begin{figure}[h]
\centering
\includegraphics[width=\columnwidth]{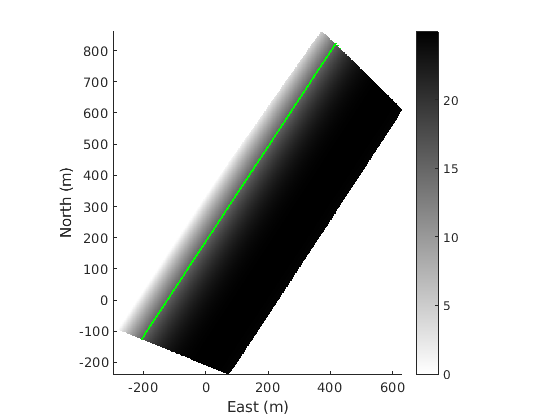}
\caption{Prior mean depth map with 15 meter isobath shown as a green line.}
\label{fig.prior}
\end{figure}

\begin{figure}[h]
\centering
\includegraphics[width=\columnwidth]{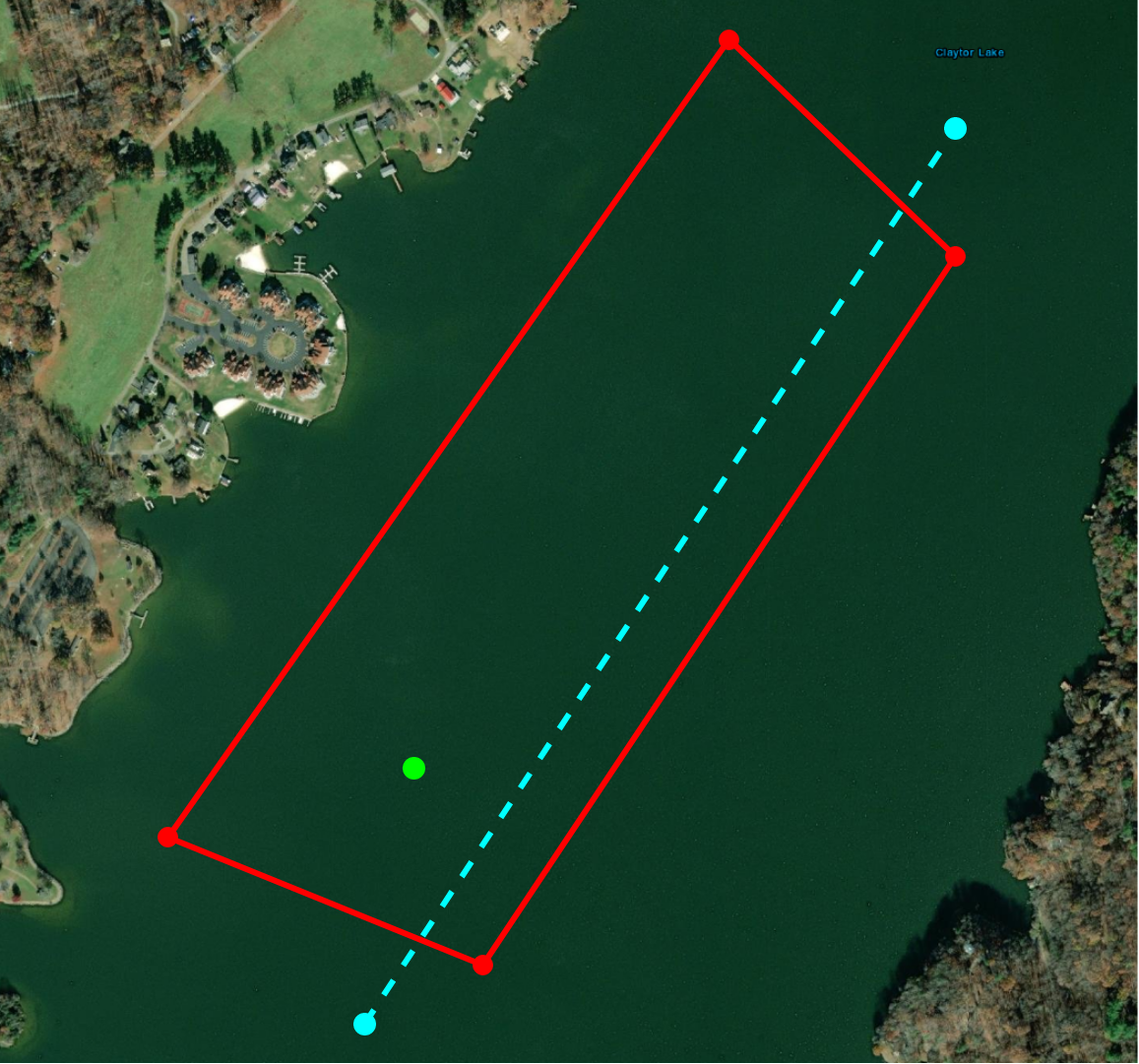}
\caption{Claytor lake operational area.}
\label{fig.search_area}
\end{figure}

The prior mean $m(p)$ is defined using a line $\mathscr{L}$. Let $d(p,\mathscr{L})$ be the minimum euclidean distance from $p$ to $\mathscr{L}$. Then the prior mean at a point $p$ is given as
\begin{equation}
    m(p) = \max(d_{max}(1 - [d(p,\mathscr{L})/c]^2), 0).
\end{equation}
Intuitively, the line $\mathscr{L}$ gives the anticipated deepest set of points in the the lake with the anticpated depth decreasing proportionally to the square of the distance from the line until a depth of zero is reached. The coefficients $c$ and $d_{max}$ are chosen to be 300 and 25 meters respectively. The specific line chosen for these trials is shown as the cyan dashed line in Figure \ref{fig.search_area}. Figure \ref{fig.prior} shows the 15 meter isobath of the prior mean function within the bounded search area. This choice of prior mean function provides a naive initial estimate of the location of the 15 meter isobath which facilitates quickly finding the true isobath location, but is not meant to represent the bathymetry accurately beyond an initial naive estimate. The parameters of the squared exponential prior covariance function in equation \eqref{eq.se} are chosen to be $\sigma_f = 5.0$, $l_c = 40.0$ meters, and the sensor measurement noise $\sigma_n = 2.5$ meters. The data retention threshold $\delta_f$ and local GP diameter $\delta_c$ presented in Section \ref{sec.gpr} are chosen to be $\delta_f = 6$ meters and $\delta_c = 40$ meters or $0.15 l_c$  and $l_c$ respectively.

\subsection{FEASIBLE PATHS}

Path planning for the vehicle is performed in the NED frame at a fixed depth of 2 meters. In planning, the vehicle state is given as $s = \{x_h, p_n, p_e \}$ where $x_h$ gives the heading in radians and $p_n$ and $p_e$ are the North and East components of the position of the state in meters from the NED origin. The set of next states available from any current state are defined by the following discrete dynamics
\begin{align}
    x_h[t+1] &= x_h[t] + a \\
    p_n[t+1] &= p_n[t] + d_m \cos(x_h[t] + a) \\
    p_e[t+1] &= p_e[t] + d_m \sin(x_h[t] + a)
\end{align}
where $a \in \mathcal{A}$ is an input angle in radians and $d_m$ is the desired euclidean distance in meters between path states. Throughout these experiments, $d_m = 30$ meters and $\mathcal{A} = \{-90, -45, -15, 0.0, 15, 45, 90 \}$ with the values of $A$ given in degrees for convenience. Let $p[t] = [p_n[t], p_e[t]]^\top$. To traverse a path, the 690 vehicle tracks a line from $p[t]$ to $p[t+1]$ until it reaches $p[t+1]$ at which point it attempts to track the line from $p[t+1]$ to $p[t+2]$. By construction, the heading of the vehicle upon reaching a point in its path should be close to the desired heading $x_h$.

\subsection{TERMINAL REWARDS}
The value of a path using equation \eqref{eq.vn} is found by appending a pair of lawnmower paths to the end of each short $n$-length path and selecting the lawnmower path that maximizes the reward of the augmented path. Lawnmower paths are given as a sequence of straight lines parallel to the heading of the final state of the short $n$-length path and continuing outward on a single side at fixed intervals in the direction of the first turn required to remain within the operational area. Notably, the value of the lawnmower path with the highest expected reward provides a lower bound on the performance of the receding horizon path. One may reasonably expect that constantly maximizing the expected reward would generate a better solution than the naive lawnmower path, however, the formal guarantee only provides that the receding horizon path will perform no worse than the naive lawnmower path. In practice, the receding horizon path generally produces a significantly improved solution.

\subsection{COMPUTATION OF REWARD ATTAINED}
Note that equation \eqref{eq.reward} assumes that a single measurement will be recorded precisely at the location contained in each path state when, in reality, measurements are taken periodically by the DVL at a rate of 10Hz with a subset of those measurements being retained according to the methods described in Section \ref{sec.gpr}. In an attempt to ensure that several measurements of the lake depth are taken near each point of interest, we consider that a location has been sampled 3 seconds after the vehicle reaches the associated waypoint. The realized reduction in ambiguity is computed as the difference between the ambiguity at the point when first beginning the transition to the point and the ambiguity at the point 3 seconds after reaching the point. The choice of 3 seconds is based on the 690 vehicles average velocity of approximately 1.6 m/s.

\subsection{RECEDING HORIZON PATH PLANNING AND EXECUTION}

\begin{figure*}[t!]
\centering
\begin{subfigure}{\columnwidth}
    \centering
    \includegraphics[width=\columnwidth]{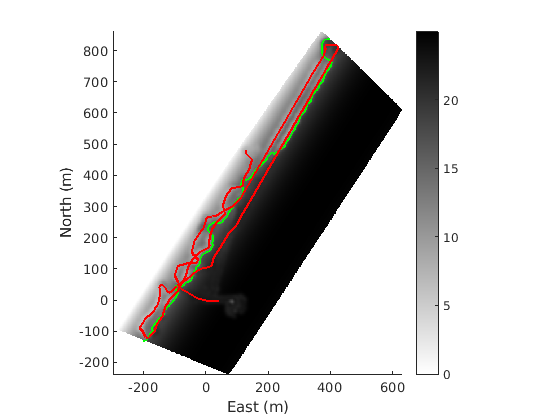}
    \caption{}
    \label{fig.rh_tc_path}
\end{subfigure}%
\begin{subfigure}{\columnwidth}
    \centering
    \includegraphics[width=\columnwidth]{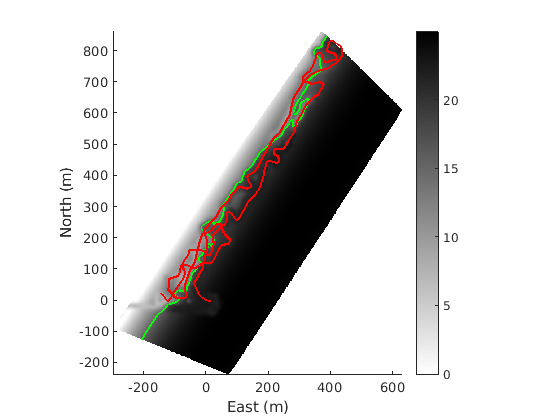}
    \caption{}
    \label{fig.rh_path}
\end{subfigure}%
\caption{The posterior predictive depth map within the bounded search region given the measurements acquired while executing the receding horizon path shown in red produced using terminal rewards (a) and the receding horizon path produced without the addition of terminal rewards (b). The 15 meter isobath is shown in green.}
\label{fig.path_results}
\end{figure*}

In this work, we utilize a basic implementation of Monte Carlo Tree Search (MCTS) \cite{browne2012survey, kocsis2006bandit, kocsis2006improved} with minor modifications to ensure that the lower bound requirements of Proposition \ref{prp_better_than_prev} are satisfied. First, the reward term of the UCT is normalized using the value of the highest-reward rollout path found for the current planning epoch and is represented as $\tilde{\mathcal{V}}_l(s_k)$. Second, instead of back-propagating the average reward of a Monte Carlo simulation, we back-propagate the maximum $\mathcal{V}_l(s_k)$ such that
\begin{equation}\label{eq.UCT}
       \mathbb{UCT} =  \frac{\mathcal{V}_l(s_k)}{ \tilde{\mathcal{V}}_l(s_k)} + C \, \sqrt{\frac{\mathrm{ln}(N)}{n}}.
\end{equation}
Lastly, at the beginning of each planning epoch, the recomputed value of the remainder of the best path found in the previous planning epoch is compared against the value of the highest-reward lawnmower path starting from the root state. The path with the highest reward is saved as the initial highest-reward rollout path for the planning epoch such that the MCTS planner is guaranteed to satisfy the critera of Proposition \ref{prp_better_than_prev} at each step. Furthermore, the comparison of the remainder of the path at each planning step against the best lawnmower path from the same initial state ensures that the guarantee of Proposition \ref{prp_better_than_prev} is extended to 
 \begin{equation}
     J(\tilde{\gamma}_l(s_k)) \geq B_l(s_k)
 \end{equation}
 at each step $k$ such that the theoretical lower bound for the receding horizon path produced is 
 \begin{equation}\label{eq.best_lb}
     J(\tilde{\gamma}_l(s_0)) \geq \underset{k \in \{1, \ldots, l \}}{\arg \max}[J(\tilde{\gamma}_k(s_0)) + B_l(s_k)]
 \end{equation}
 as discussed in Section 7 of \cite{biggs2021multi}.

 Defining a step to be a line segment connecting consecutive states in a path. The planning horizon is $5$ steps, the execution horizon is a single step, and the total mission length $l$ is 100 steps. For comparison, one path was planned using the proposed receding horizon method with terminal rewards and another receding horizon was planned without the addition of terminal rewards. The receding horizon path produced using the proposed method is shown as the solid red line in Figure \ref{fig.rh_tc_path}. The receding horizon path produced without the use of terminal rewards is similarly shown in Figure \ref{fig.rh_path}.

\begin{figure}[b!]
\centering
\includegraphics[width=\columnwidth]{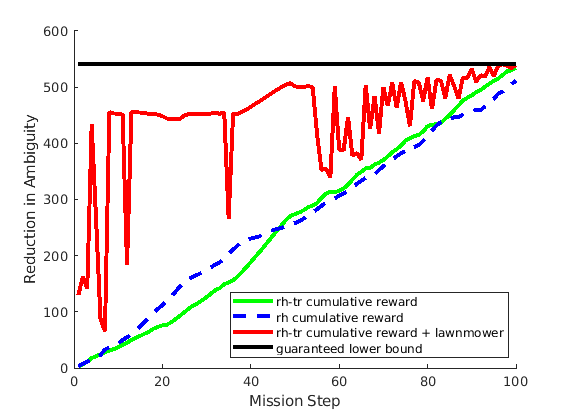}
\caption{Cumulative rewards for the RH path and RH path constructed with terminal rewards are shown in dashed blue and solid green respectively. The cumulative reward for the RH path constructed with terminal rewards plus the lower bound on the value-to-go is shown as the solid red line. The guaranteed lower bound is shown as the solid black line.}
\label{fig.results}
\end{figure}

\section{RESULTS AND CONCLUSIONS}\label{sec.conclusions}
The results of planning and executing receding horizon paths are shown in Figures \ref{fig.path_results} and \ref{fig.results}. Figure \ref{fig.path_results} shows the posterior depth maps produced using equations \eqref{eq.pred_mean} and \eqref{eq.pred_var} and the data sets acquired while traversing the receding horizon paths according to the methods discussed in Section \ref{sec.gpr}. Figure \ref{fig.path_results} also shows the 15 meter isobath in green. The cumulative reward for the receding horizon path produced without the use of terminal rewards is shown as the dashed blue line in Figure \ref{fig.results} with the cumulative reward for the receding horizon path produced with terminal rewards shown as the solid green line. The solid black line shows the value of the lower bound given in equation \ref{eq.best_lb} which is the maximum value of the solid red line. The red line gives the sum of the current cumulative reward attained by the receding horizon path up to step $k$ plus the lower bound on the value-to-go.

Notably, Figure \ref{fig.results} shows that both receding horizon paths performed significantly better than the value of the initial lower bound $B_l(s_0)$ with the receding horizon path produced using the proposed methods performing slightly better than the receding horizon path that did not include terminal rewards. While there is no guarantee that paths produced using the proposed methods will perform better or worse than receding horizon paths produced using other methods, these results do not indicate any degradation in performance for including terminal rewards defined by naive heuristic paths. Furthermore, the results of these experiments appear to further confirm the results of Proposition \ref{prp_better_than_prev}.

\IEEEtriggeratref{10}
\bibliography{sources}
\bibliographystyle{ieeetr}


\end{document}